\documentclass[letterpaper]{article}
\usepackage[preprint]{aaai2027}
\usepackage[hyphens]{url}
\usepackage{graphicx}
\usepackage{natbib}
\usepackage{caption}
\usepackage{algorithm}
\usepackage{algorithmic}

\usepackage{booktabs}
\usepackage{multirow}
\usepackage{placeins}

\usepackage{amsmath}
\usepackage{amssymb}

\newcommand{\R}{\mathbb{R}}

\newcommand{\simplex}{\Delta}

\newcommand{\Lnce}{\mathcal{L}_{\mathrm{nce}}}
\newcommand{\Lkl}{\mathcal{L}_{\mathrm{kl}}}
\newcommand{\Linst}{\mathcal{L}_{\mathrm{inst}}}
\newcommand{\Ldiv}{\mathcal{L}_{\mathrm{div}}}

\newcommand{\ze}{z_e}
\newcommand{\zv}{z_v}
\newcommand{\zhate}{\hat{z}_e}
\newcommand{\zhatv}{\hat{z}_v}
\newcommand{\alphae}{\alpha_e}
\newcommand{\alphav}{\alpha_v}

\title{ProtoEEG: Structured Prototype Alignment for EEG Visual Decoding}

\author{Kieren Yu, Chang Huang, Ziyang Liu, Kaishun Wu}
\affiliations{}

\begin{document}

\maketitle
\pagestyle{plain}
\thispagestyle{plain}

\begin{abstract}
EEG visual decoding remains difficult because scalp recordings are noisy, spatially coarse, and variable across trials and subjects.
Most systems align each trial with a dense image embedding, although that embedding mixes semantic content with instance-specific texture, pose, and background.
This target can demand details absent from EEG and provides no explicit account of the visual factors supporting a prediction.
We propose \textbf{ProtoEEG}, a prototype-mediated framework that reformulates EEG--image alignment as soft composition over a shared learnable visual codebook.
A trainable EEG encoder and a frozen visual teacher produce assignment distributions over the same $K=64$ prototypes.
Their weighted reconstructions form a structured representation for retrieval and image generation.
Prototype contrast supplies the main cross-modal training signal, distribution alignment coordinates codebook use across modalities, an instance anchor preserves fine-grained discrimination, and diversity regularization prevents redundant prototypes.
The shared assignment profile also makes the learned bottleneck directly testable through prototype retention, random selection, and identity shuffling.
Across five seeds, ProtoEEG reaches 64.0\% Top-1 and 91.5\% Top-5 on THINGS-EEG2 200-way retrieval.
After retraining on EEG-ImageNet, it obtains 44.8\% Top-1 and 73.8\% Top-5 in 80-way retrieval.
ProtoEEG also ranks first on all seven reconstruction metrics.
Removing prototype contrast lowers Top-1 by 56.9 percentage points, while shuffling prototype identities reduces it to 0.5\%.
Prototype analyses further show distributed assignments and broad codebook use rather than one-prototype routing.
These results support structured alignment targets as a practical alternative to increasing encoder or teacher capacity alone.
\end{abstract}

\section{Introduction}

Visual decoding has been studied across several neuroimaging modalities. Functional magnetic resonance imaging (fMRI) can reconstruct viewed images from blood-oxygen-level-dependent signals~\citep{benchetrit2024braindecoding,scotti2023mindeye}, but it requires expensive scanners and restricts subject movement. Electroencephalography (EEG) is portable, inexpensive, and records neural dynamics at millisecond resolution~\citep{spampinato2017}. Its scalp measurements are spatially coarse and vary across trials and people. A decoder must therefore recover visual distinctions without assuming that texture, viewpoint, or background is recorded consistently.

\begin{figure}[t]
  \centering
  \includegraphics[width=\columnwidth]{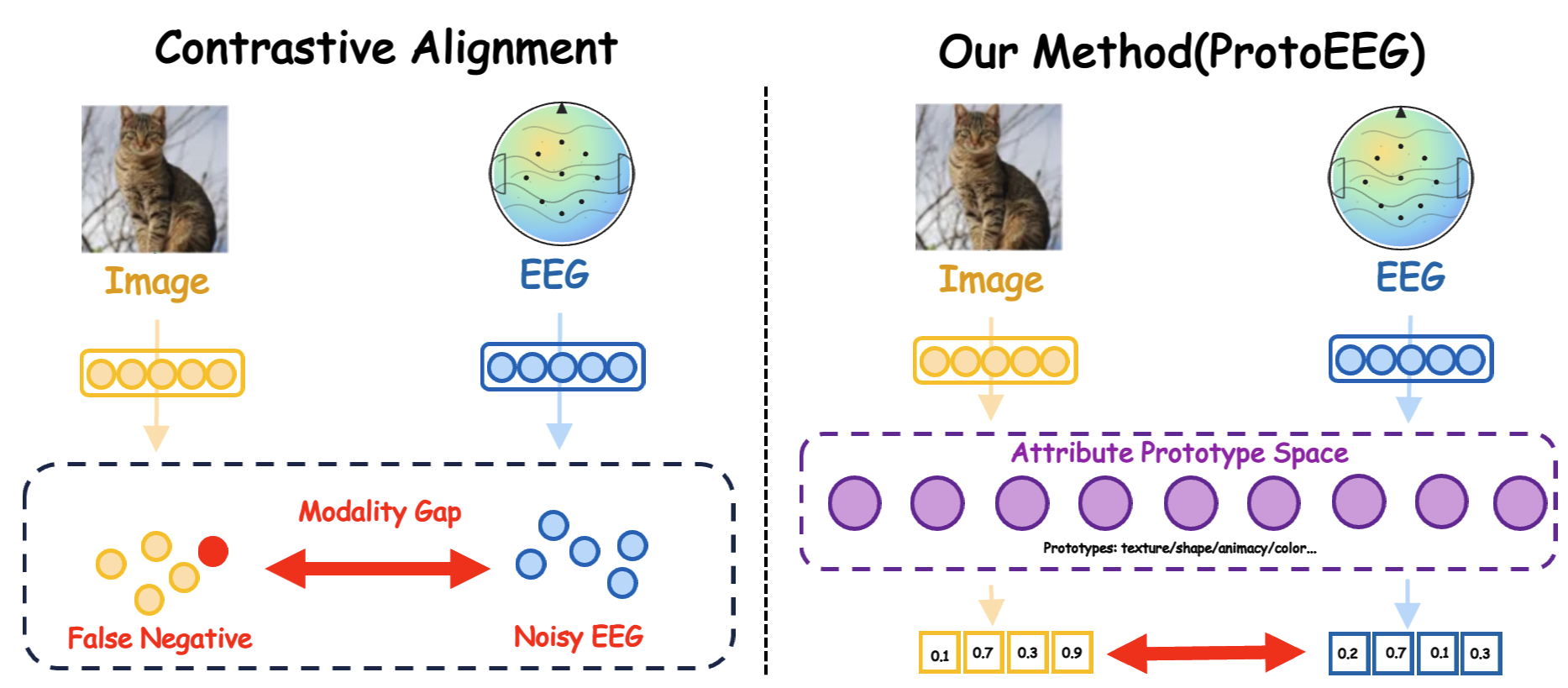}
  \caption{Dense alignment matches EEG to instance-specific image embeddings and can treat related concepts as negatives, as shown on the left. ProtoEEG instead maps both modalities to soft assignments over $K=64$ shared visual prototypes, shown on the right.}
  \label{fig:intro}
\end{figure}

Deep learning has moved EEG visual decoding from category prediction toward representation alignment and image generation. Early approaches learned joint EEG--image embeddings for classification~\citep{spampinato2017,palazzo2020}. Recent methods align EEG with pretrained visual features~\citep{song2024nice,li2024atm,zhang2025cognitioncapturer,li2025brainflora,liu2025vieeg,singh2024robust}, modify visual supervision~\citep{wu2025ubp,wu2026ats,zhang2026neurobridge}, or condition generative models on EEG embeddings~\citep{bai2024dreamdiffusion,dai2026avde}. Large-scale pretraining on EEG has also shown promise for learning generic representations~\citep{jiang2024large,yupatchcode,zhou2026csbrain,wang2025cbramod,el2026reve}.
These lines of work improve temporal modeling, visual supervision, or generation, yet they primarily change the predictor or teacher rather than asking whether a dense visual embedding is an appropriate target for a noisy neural measurement.

Most of these methods retain the same assumption: an EEG trial should match one dense, instance-level image representation~\citep{song2024nice,li2024atm,wu2025ubp,wu2026ats,zhang2026neurobridge}. Such a representation combines category and coarse shape with viewpoint, texture, lighting, background, and spatial arrangement, although EEG may support only some of these factors. Direct alignment can penalize the neural encoder for variation absent from the measurement and turn related images into negatives. The cross-modal gap adds another mismatch~\citep{liang2022mindthegap}. The choice of alignment target is therefore as important as encoder capacity.

The alignment target should remain tied to a visual teacher, represent uncertainty compositionally, and expose that composition for analysis. Prototype methods organize data around reusable representatives~\citep{snell2017prototypical,caron2018deepcluster,caron2020swav,li2021pcl}. Hard assignments are too restrictive because one EEG trial may not support a single categorical choice. We instead use a learnable codebook with soft assignments. The prototypes define shared visual directions, while each sample assigns graded weights to them.

We propose \textbf{ProtoEEG}, which aligns EEG and images through a shared visual prototype space. Figure~\ref{fig:intro} shows the main design. A trainable EEG encoder and a frozen visual teacher produce soft assignments over the same $K=64$ codebook, then reconstruct assignment-weighted embeddings for retrieval and image generation. Prototype contrast aligns the decoded representation, while distribution, instance, and diversity terms coordinate codebook use and prevent collapse. The code supports retrieval, reconstruction, and controlled perturbations without assigning neurocognitive meanings to individual prototypes.

Our main contributions are:
\begin{itemize}
    \item We propose ProtoEEG, a prototype-alignment framework built around a shared learnable codebook. Soft assignments express graded EEG evidence without forcing one prototype per trial.
    \item We combine prototype contrast for cross-modal discrimination, distribution alignment for shared codebook use, an instance anchor for fine-grained distinctions, and diversity regularization to prevent redundant prototypes.
    \item Experiments across retrieval and reconstruction show consistent gains across datasets and visual teachers. ProtoEEG reaches 64.0\% Top-1 and 91.5\% Top-5 on THINGS-EEG2, obtains 44.8\% Top-1 on EEG-ImageNet after retraining, and leads reconstruction metrics.
\end{itemize}

\section{Related Work}

\paragraph{EEG visual retrieval and representation alignment.}
Early studies decoded object categories from single-trial EEG through joint EEG--image embeddings~\citep{spampinato2017,palazzo2020}. Recent retrieval systems use contrastive learning to align neural signals with visual representations~\citep{song2024nice,li2024atm,zhang2025cognitioncapturer,li2025brainflora,liu2025vieeg,fu2025brainvis,singh2024robust}. Other approaches modify visual supervision through uncertainty-aware blur priors, adaptive teacher compression, or bidirectional cognitive alignment~\citep{wu2025ubp,wu2026ats,zhang2026neurobridge,rajabi2025humanaligned}. These changes address signal extraction, target quality, and robustness. They still rely mainly on dense instance-level alignment and ask EEG to match fine-grained image features. ProtoEEG instead changes the coordinates used for cross-modal comparison.

\paragraph{EEG-to-image reconstruction and generative priors.}
DreamDiffusion and ATM condition diffusion models on EEG-derived representations~\citep{bai2024dreamdiffusion,li2024atm}; AVDE uses autoregressive decoding~\citep{dai2026avde}. Related fMRI and MEG systems combine contrastive alignment with diffusion priors~\citep{benchetrit2024braindecoding,scotti2023mindeye,scotti2024mindeye2,ozcelik2023braindiffuser}. Generative priors improve synthesis, but the EEG condition remains a dense vector with no explicit decomposition. Reconstruction quality can therefore improve without showing which part of the neural representation supplies category, shape, or appearance information. Our reconstruction experiment uses the prototype code learned for retrieval, rather than a separate representation trained only for generation.

\paragraph{Prototype learning and multimodal alignment.}
Prototype-based representations include few-shot class centroids~\citep{snell2017prototypical}, cluster codebooks for self-supervised learning~\citep{caron2018deepcluster,caron2020swav,li2021pcl}, and interpretable part prototypes~\citep{chen2019protopnet}. Multimodal variants align prototypes across images and text or clinical reports~\citep{cheng2023prior,qin2023capro}, providing structure across modalities with different feature distributions~\citep{liang2022mindthegap}. Existing methods either use hard assignments that are poorly suited to noisy EEG or keep prototypes within one modality. Neither setting addresses soft neural--visual alignment. ProtoEEG shares one trainable codebook across EEG and images and retains the resulting assignment profile at inference.

\begin{figure*}[t]
  \centering
  \includegraphics[width=\textwidth]{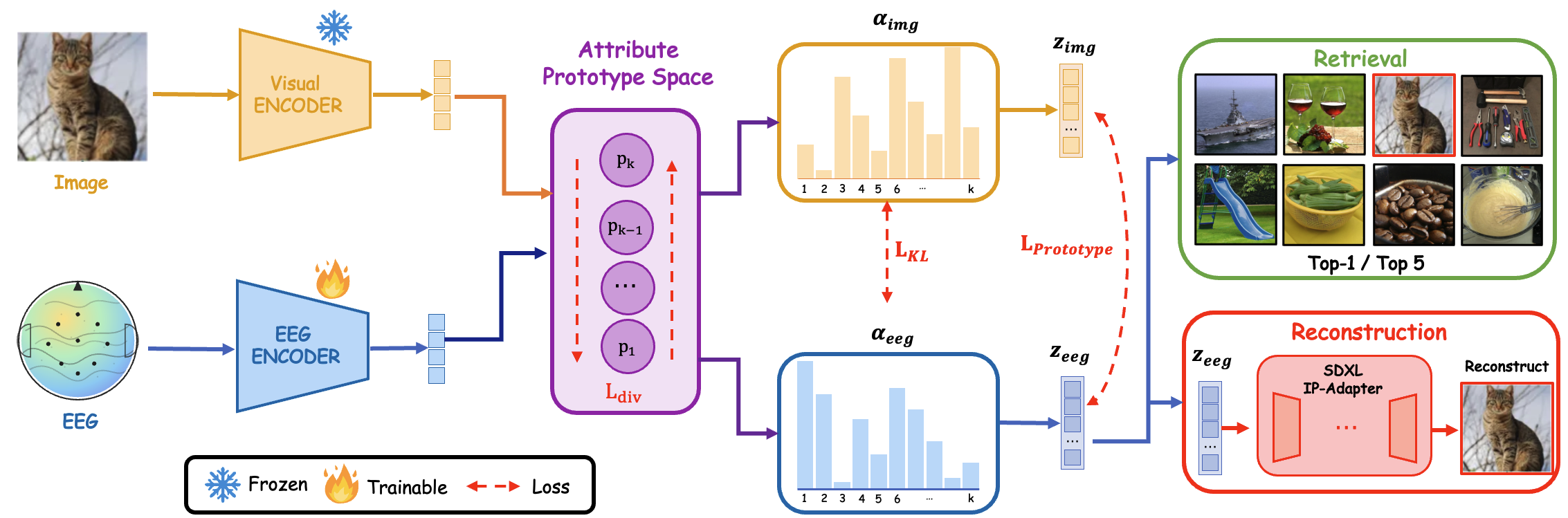}
  \caption{Overview of ProtoEEG. A frozen visual teacher and a trainable EEG encoder produce soft assignments over a shared prototype codebook. The diagram denotes $\zhatv$ and $\zhate$ by $z_{\mathrm{img}}$ and $z_{\mathrm{eeg}}$. These embeddings support retrieval and image generation. Training uses prototype contrast $\Lnce$, distribution alignment $\Lkl$, and diversity regularization $\Ldiv$; the instance anchor $\Linst$.}
  \label{fig:overview}
\end{figure*}

\section{Method}

\subsection{Overview and Problem Definition}

Figure~\ref{fig:overview} shows the complete architecture. A trainable EEG encoder and projection head produce the normalized embedding $\ze = g_e(f_e(X)) \in \R^d$. A visual encoder and projection head similarly produce $\zv = g_v(f_v(I)) \in \R^d$. We freeze the visual backbone $f_v$ and train only $f_e$, $g_e$, $g_v$, and the prototype codebook. Freezing $f_v$ fixes the visual coordinate system throughout training so the EEG encoder learns to use this space rather than changing the teacher to follow EEG noise. Both modalities produce soft assignments over the same prototype matrix, and their assignment-weighted reconstructions replace direct recovery of dense visual embeddings.

Given paired data $\mathcal{D} = \{(X_i, I_i, y_i)\}_{i=1}^{N}$, $X_i \in \R^{C \times T}$ denotes a single-trial EEG recording, $I_i$ its stimulus image, and $y_i$ its concept label. Labels are used for evaluation, not for training the codebook. Our goal is to map EEG and images into a common representation for retrieval and reconstruction.

\subsection{Shared Prototype Representation}

\paragraph{Prototype codebook.}
The shared visual basis is a learnable matrix $P = [p_1, \ldots, p_K]^\top \in \R^{K \times d}$ containing $K$ unit-norm prototypes. We initialize them from random training-image embeddings and update them through paired EEG--image supervision. They are neither class centroids nor pre-labelled semantic concepts. The same matrix $P$ is shared across all trials, subjects, and images within a trained model, providing stable coordinates for cross-modal comparison, while the assignment vector changes with each observation and records which parts of that basis are supported. This distinction separates dataset-level structure from trial-level evidence: learning $P$ across samples gives recurring visual patterns a shared coordinate, while $\alpha_m$ records how one observation uses that coordinate.

\paragraph{Soft assignment.}
For a normalized embedding $z_m$ from modality $m \in \{e, v\}$, the soft assignment over prototypes is
\begin{equation}
  \alpha_m = \mathrm{softmax}\!\left(\frac{z_m P^\top}{\tau}\right) \in \simplex^{K-1},
\label{eq:assignment}
\end{equation}
where $\tau > 0$ controls assignment sharpness and $\simplex^{K-1}$ is the probability simplex. EEG noise makes a unique prototype assignment unreliable. Soft assignments allow several visual factors to explain one trial and represent evidence as graded weights. Because $\alphae$ and $\alphav$ use the same codebook, their coordinates can be aligned directly.

The temperature $\tau$ determines the effective resolution of this description. A low value concentrates probability on a small number of prototypes, approaching hard routing, whereas a higher value distributes support more broadly. Retaining a differentiable full assignment lets gradients jointly refine prototype identities and mixture weights. It also creates a controlled interface for later analysis: entries can be retained, removed, randomized, or permuted without changing the underlying EEG encoder.

We use the same temperature and prototype matrix for both modalities. Any difference between $\alphae$ and $\alphav$ then reflects the encoded observations rather than a modality-specific codebook. The distribution loss below reduces this difference for paired samples without requiring their raw embeddings to be identical.

\paragraph{Prototype reconstruction.}
The prototype-reconstructed embedding is
\begin{equation}
  \hat{z}_m = \alpha_m P = \sum_{k=1}^{K} \alpha_m^{(k)}\, p_k \in \R^d.
\label{eq:reconstruction}
\end{equation}
For EEG, $\zhate$ is the representation used for decoding. For gallery images, $\zhatv$ serves as the comparison target. Both lie in the same codebook span. The prototype bottleneck retains shared visual structure while constraining representations away from unstable instance-specific details that EEG cannot encode. The resulting representations pool evidence across visually similar stimuli, as images with similar content naturally receive similar assignment profiles.

The reconstruction is deterministic once $P$ and $\alpha_m$ are fixed. Retrieval scores can therefore be traced to an ordered set of prototype weights, and perturbing a weight changes the same representation used by the decoder. This link is the basis of the assignment interventions reported in Section 4.

Because the weights are nonnegative and sum to one, each reconstruction lies in the convex hull of the learned prototypes. This constraint is weaker than assigning one class but more structured than predicting an unrestricted vector. An individual prototype is not a complete semantic concept. Information is carried by the ordered profile and interactions among several basis vectors. We interpret the model by testing this decomposition rather than naming prototypes after training.

\subsection{Learning Objective}

The model is trained with four complementary losses. The prototype contrastive loss aligns prototype-reconstructed EEG and visual embeddings via symmetric InfoNCE~\citep{oord2018infonce}:
\begin{equation}
  \Lnce = -\frac{1}{2B}\sum_{i=1}^{B}\!\left[
    \log\frac{e^{\zhate^i \cdot \zv^i / \tau'}}{\sum_j e^{\zhate^i \cdot \zv^j / \tau'}}
    + \log\frac{e^{\zv^i \cdot \zhate^i / \tau'}}{\sum_j e^{\zv^i \cdot \zhate^j / \tau'}}
  \right],
\label{eq:nce}
\end{equation}
where $B$ is the batch size and $\tau'$ is learned. This loss organizes the decoded representation for cross-modal discrimination. Its gradients update both prototype assignment and reconstruction. Without it, paired samples may reach similar embeddings through incompatible assignments.

A distribution alignment term encourages both modalities to use the codebook similarly for the same stimulus:
\begin{equation}
  \Lkl = \frac{1}{2B}\sum_{i=1}^{B}\!\left[
    D_{\mathrm{KL}}(\alphae^i \| \alphav^i) + D_{\mathrm{KL}}(\alphav^i \| \alphae^i)
  \right].
\label{eq:kl}
\end{equation}
This term aligns the assignment profiles of paired samples and prevents them from reaching similar reconstructions through incompatible prototype weights.

An instance-level anchor operates on raw embeddings to preserve discriminability:
\begin{equation}
  \Linst = -\frac{1}{2B}\sum_{i=1}^{B}\!\left[
    \log\frac{e^{\ze^i \cdot \zv^i / \tau'}}{\sum_j e^{\ze^i \cdot \zv^j / \tau'}}
    + \log\frac{e^{\zv^i \cdot \ze^i / \tau'}}{\sum_j e^{\zv^i \cdot \ze^j / \tau'}}
  \right].
\label{eq:inst}
\end{equation}
This preserves distinctions that prototype reconstruction may smooth. Without it, the bottleneck can discard fine-grained details needed for candidate ranking.

To prevent prototype collapse, we penalize pairwise cosine similarity:
\begin{equation}
  \Ldiv = \frac{1}{K(K-1)}\sum_{j \neq k} p_j \cdot p_k.
\label{eq:div}
\end{equation}
This discourages redundant basis vectors. Without it, several rows of $P$ can converge to similar directions and reduce the effective codebook size. Table~\ref{tab:ablation} tests each term separately.

The total loss is
\begin{equation}
  \mathcal{L} = \Lnce + \lambda_{\mathrm{kl}}\,\Lkl + \lambda_{\mathrm{inst}}\,\Linst + \lambda_{\mathrm{div}}\,\Ldiv.
\label{eq:total}
\end{equation}

Optimization updates the EEG encoder, both projection heads, and the prototype matrix jointly, while the visual backbone remains fixed. After each update, prototype rows are normalized again so that assignment scores continue to reflect angular similarity rather than changing vector magnitude. Initialization from training-image embeddings gives the codebook a visual starting geometry, but its final organization is determined by the paired objective because gradients from both assignments and decoded representations reach $P$. The complete training procedure is summarized in Algorithm~\ref{alg:protoeeg}.

\begin{algorithm}[t]
\caption{ProtoEEG Training}
\label{alg:protoeeg}
\begin{algorithmic}[1]
\REQUIRE Dataset $\mathcal{D} = \{(X_i, I_i)\}_{i=1}^N$, frozen visual encoder $f_v$, codebook size $K$, temperature $\tau$
\ENSURE Trained EEG encoder $f_e$, projections $g_e, g_v$, prototype matrix $P$
\STATE Initialize $P$ from $K$ random image embeddings; L2-normalize rows
\FOR{each training batch $(X, I)$}
  \STATE Encode: $\zv \gets g_v(f_v(I))/\|g_v(f_v(I))\|_2$, $\ze \gets g_e(f_e(X))/\|g_e(f_e(X))\|_2$
  \STATE Assign: $\alphav \gets \mathrm{softmax}(\zv P^\top / \tau)$, $\alphae \gets \mathrm{softmax}(\ze P^\top / \tau)$
  \STATE Reconstruct: $\zhate \gets \alphae P$
  \STATE Compute $\Lnce$, $\Lkl$, $\Linst$, and $\Ldiv$ using Eqs.~\ref{eq:nce}--\ref{eq:div}
  \STATE Update $f_e, g_e, g_v, P$ on the objective in Eq.~\ref{eq:total}
  \STATE L2-normalize each row of $P$
\ENDFOR
\end{algorithmic}
\end{algorithm}

% Schedule the main retrieval table early so the two-column float appears
% on the first experiments page; its discussion remains in Section 4.
\begin{table*}[!t]
\centering
\small
\setlength{\tabcolsep}{3.6pt}
\renewcommand{\arraystretch}{0.95}
\begin{tabular*}{\textwidth}{@{\extracolsep{\fill}}l*{16}{c}@{}}
\toprule
\multicolumn{17}{c}{\textbf{(a) Intra-subject: train and test on one subject}} \\
\addlinespace[1pt]
& \multicolumn{8}{c}{\textbf{CLIP family}} & \multicolumn{8}{c}{\textbf{RN50 family}} \\
\cmidrule(lr){2-9}\cmidrule(lr){10-17}
& \multicolumn{2}{c}{\textbf{NICE}} & \multicolumn{2}{c}{\textbf{ATM}} & \multicolumn{2}{c}{\textbf{AVDE}} & \multicolumn{2}{c}{\textbf{ProtoEEG}} & \multicolumn{2}{c}{\textbf{UBP}} & \multicolumn{2}{c}{\textbf{ATS}} & \multicolumn{2}{c}{\textbf{NeuroBridge}} & \multicolumn{2}{c}{\textbf{ProtoEEG}} \\
\cmidrule(lr){2-3}\cmidrule(lr){4-5}\cmidrule(lr){6-7}\cmidrule(lr){8-9}\cmidrule(lr){10-11}\cmidrule(lr){12-13}\cmidrule(lr){14-15}\cmidrule(lr){16-17}
\textbf{Subject} & \textbf{T1} & \textbf{T5} & \textbf{T1} & \textbf{T5} & \textbf{T1} & \textbf{T5} & \textbf{T1} & \textbf{T5} & \textbf{T1} & \textbf{T5} & \textbf{T1} & \textbf{T5} & \textbf{T1} & \textbf{T5} & \textbf{T1} & \textbf{T5} \\
\midrule
Sub-01 & 20.1 & 47.9 & 23.2 & 51.2 & 25.0 & 55.2 & \textbf{39.3} & \textbf{71.8} & 41.2 & 70.5 & 53.0 & 79.0 & 50.0 & 77.6 & \textbf{62.9} & \textbf{89.7} \\
Sub-02 & 19.2 & 36.9 & 18.8 & 43.2 & 24.1 & 51.0 & \textbf{32.7} & \textbf{63.5} & 51.2 & 80.9 & 62.0 & 87.5 & \textbf{63.2} & 90.6 & 61.8 & \textbf{94.1} \\
Sub-03 & 21.2 & 53.8 & 27.3 & 57.0 & 27.5 & 58.6 & \textbf{35.5} & \textbf{66.6} & 51.2 & 82.0 & 61.5 & 89.0 & 61.6 & 91.1 & \textbf{64.9} & \textbf{93.6} \\
Sub-04 & 22.4 & 50.4 & 28.0 & 54.1 & 29.8 & 54.7 & \textbf{42.1} & \textbf{73.9} & 51.1 & 76.9 & 57.0 & 86.5 & \textbf{61.4} & \textbf{90.0} & 58.5 & 88.5 \\
Sub-05 & 14.4 & 31.6 & 16.8 & 39.5 & 25.4 & 50.3 & \textbf{26.4} & \textbf{53.5} & 42.2 & 72.8 & \textbf{55.0} & 84.0 & 54.8 & \textbf{85.0} & 51.4 & 81.9 \\
Sub-06 & 26.1 & 56.3 & 28.0 & 59.2 & 33.5 & 60.3 & \textbf{36.1} & \textbf{66.4} & 57.5 & 83.5 & 68.0 & 90.5 & \textbf{69.7} & 92.9 & 67.2 & \textbf{96.3} \\
Sub-07 & 26.9 & 55.7 & 26.8 & 53.7 & 27.4 & 55.2 & \textbf{35.2} & \textbf{69.0} & 49.0 & 79.9 & 53.0 & 84.0 & 62.7 & 88.8 & \textbf{64.5} & \textbf{93.6} \\
Sub-08 & 38.2 & 67.4 & 39.3 & 71.5 & 41.7 & 71.3 & \textbf{47.9} & \textbf{80.7} & 58.6 & 85.8 & 66.5 & 91.0 & \textbf{71.2} & \textbf{95.1} & 69.4 & 92.7 \\
Sub-09 & 23.4 & 53.2 & 24.5 & 51.2 & 26.1 & 52.1 & \textbf{35.7} & \textbf{63.4} & 45.1 & 76.2 & 58.5 & 86.0 & 64.0 & 91.0 & \textbf{69.7} & \textbf{92.0} \\
Sub-10 & 29.8 & 58.6 & 37.2 & 67.7 & 39.5 & 73.0 & \textbf{45.6} & \textbf{79.8} & 61.5 & 88.2 & 67.5 & 89.0 & \textbf{73.6} & \textbf{97.1} & 69.4 & 92.9 \\
\midrule
\textbf{Avg} & 24.2 & 51.2 & 27.0 & 54.8 & 30.0 & 58.2 & \textbf{37.7} & \textbf{68.9} & 50.9 & 79.7 & 60.2 & 86.7 & 63.2 & 89.9 & \textbf{64.0} & \textbf{91.5} \\
\bottomrule
\end{tabular*}

\vspace{0.55em}

\begin{tabular*}{\textwidth}{@{\extracolsep{\fill}}l*{16}{c}@{}}
\toprule
\multicolumn{17}{c}{\textbf{(b) Inter-subject: leave one subject out for test}} \\
\addlinespace[1pt]
& \multicolumn{8}{c}{\textbf{CLIP family}} & \multicolumn{8}{c}{\textbf{RN50 family}} \\
\cmidrule(lr){2-9}\cmidrule(lr){10-17}
& \multicolumn{2}{c}{\textbf{NICE}} & \multicolumn{2}{c}{\textbf{ATM}} & \multicolumn{2}{c}{\textbf{AVDE}} & \multicolumn{2}{c}{\textbf{ProtoEEG}} & \multicolumn{2}{c}{\textbf{UBP}} & \multicolumn{2}{c}{\textbf{ATS}} & \multicolumn{2}{c}{\textbf{NeuroBridge}} & \multicolumn{2}{c}{\textbf{ProtoEEG}} \\
\cmidrule(lr){2-3}\cmidrule(lr){4-5}\cmidrule(lr){6-7}\cmidrule(lr){8-9}\cmidrule(lr){10-11}\cmidrule(lr){12-13}\cmidrule(lr){14-15}\cmidrule(lr){16-17}
\textbf{Subject} & \textbf{T1} & \textbf{T5} & \textbf{T1} & \textbf{T5} & \textbf{T1} & \textbf{T5} & \textbf{T1} & \textbf{T5} & \textbf{T1} & \textbf{T5} & \textbf{T1} & \textbf{T5} & \textbf{T1} & \textbf{T5} & \textbf{T1} & \textbf{T5} \\
\midrule
Sub-01 & 7.6 & 22.8 & 10.5 & 26.8 & 14.1 & 32.2 & \textbf{15.8} & \textbf{39.5} & 11.5 & 29.7 & 17.0 & 37.5 & 23.2 & 52.4 & \textbf{25.8} & \textbf{55.7} \\
Sub-02 & 5.9 & 20.5 & 7.1 & 24.8 & \textbf{17.0} & 38.4 & 16.6 & \textbf{42.8} & 15.5 & 40.0 & 21.5 & 49.5 & 21.2 & 49.3 & \textbf{24.1} & \textbf{52.6} \\
Sub-03 & 6.0 & 22.3 & \textbf{11.9} & \textbf{33.8} & 9.1 & 21.8 & 9.8 & 28.4 & 9.8 & 27.0 & 7.5 & 20.0 & 13.2 & 36.5 & \textbf{14.6} & \textbf{38.9} \\
Sub-04 & 6.3 & 20.7 & 14.7 & 39.4 & 15.2 & 32.5 & \textbf{15.9} & \textbf{40.6} & 13.0 & 32.3 & 13.5 & 41.5 & 17.0 & 45.3 & \textbf{19.2} & \textbf{47.1} \\
Sub-05 & 4.4 & 18.3 & 7.0 & 23.9 & 12.5 & 32.4 & \textbf{12.9} & \textbf{35.7} & 8.8 & 33.8 & 7.0 & 27.0 & 14.5 & 37.7 & \textbf{16.3} & \textbf{40.2} \\
Sub-06 & 5.6 & 22.2 & 11.1 & 35.8 & 17.3 & 38.6 & \textbf{17.5} & \textbf{43.9} & 11.7 & 31.0 & 14.5 & 37.0 & 25.0 & 55.0 & \textbf{27.8} & \textbf{57.8} \\
Sub-07 & 5.6 & 19.7 & \textbf{16.1} & \textbf{43.5} & 7.4 & 20.4 & 10.5 & 31.2 & 10.2 & 23.8 & 10.0 & 33.5 & 15.3 & 45.1 & \textbf{18.2} & \textbf{46.9} \\
Sub-08 & 6.3 & 22.0 & 15.0 & 40.3 & 18.5 & 40.1 & \textbf{19.1} & \textbf{45.8} & 12.2 & 32.2 & 13.0 & 30.5 & 20.1 & 44.9 & \textbf{23.5} & \textbf{49.5} \\
Sub-09 & 5.7 & 17.6 & 4.9 & 22.7 & 13.2 & 33.6 & \textbf{13.8} & \textbf{37.4} & 15.5 & \textbf{40.5} & 9.5 & 28.5 & 13.7 & 36.5 & \textbf{16.4} & 39.4 \\
Sub-10 & 8.4 & 28.3 & 20.5 & 46.5 & 18.0 & 39.3 & \textbf{23.1} & \textbf{52.7} & 16.0 & 43.5 & 26.0 & 52.5 & 27.2 & \textbf{56.3} & \textbf{34.1} & 55.9 \\
\midrule
\textbf{Avg} & 6.2 & 21.4 & 11.8 & 33.7 & 14.2 & 32.9 & \textbf{15.5} & \textbf{39.8} & 12.4 & 33.4 & 14.0 & 35.8 & 19.0 & 45.9 & \textbf{22.0} & \textbf{48.4} \\
\bottomrule
\end{tabular*}
\caption{THINGS-EEG2 200-way retrieval accuracy (\%). ProtoEEG reports five-seed means. Bold marks best within each visual-encoder family. ProtoEEG consistently outperforms dense alignment baselines in both intra-subject and inter-subject settings, with CLIP gains uniform across subjects while RN50 improvements are more variable.}
\label{tab:retrieval}
\end{table*}

% Queue the second full-width result table for the following page so the
% experimental floats are distributed rather than collected on one page.
\begin{table*}[t]
\centering
\small
\setlength{\tabcolsep}{3pt}
\renewcommand{\arraystretch}{0.98}
\begin{tabular*}{\textwidth}{@{\extracolsep{\fill}}ll*{9}{c}@{}}
\toprule
\textbf{Method} & \textbf{Metric} & \textbf{Sub-01} & \textbf{Sub-02} & \textbf{Sub-03} &
\textbf{Sub-04} & \textbf{Sub-05} & \textbf{Sub-06} & \textbf{Sub-07} &
\textbf{Sub-08} & \textbf{Avg} \\
\midrule
\multirow{2}{*}{ATM}
 & Top-1 & 30.6 & 24.5 & 21.3 & 15.1 & 20.5 & 18.7 & 14.6 & 9.1 & 19.3 \\
 & Top-5 & 63.3 & 62.4 & 64.1 & 42.8 & 55.7 & 69.9 & 67.6 & 35.3 & 57.6 \\
\addlinespace[1pt]
\multirow{2}{*}{AVDE}
 & Top-1 & 30.8 & \textbf{27.0} & 22.7 & 15.4 & \textbf{21.8} & 32.9 & 30.1 & 14.4 & 24.4 \\
 & Top-5 & 63.4 & \textbf{62.8} & 64.3 & 43.5 & \textbf{56.6} & 70.3 & 69.5 & 38.8 & 58.7 \\
\addlinespace[1pt]
\multirow{2}{*}{\textbf{ProtoEEG}}
 & Top-1 & \textbf{50.2} & 16.9 & \textbf{56.5} & \textbf{25.3} & 17.7 & \textbf{57.4} & \textbf{57.3} & \textbf{77.1} & \textbf{44.8} \\
 & Top-5 & \textbf{88.6} & 51.2 & \textbf{82.3} & \textbf{48.6} & 41.4 & \textbf{93.6} & \textbf{85.5} & \textbf{98.8} & \textbf{73.8} \\
\bottomrule
\end{tabular*}
\caption{EEG-ImageNet 80-way accuracy (\%). ProtoEEG reports five-seed means.}
\label{tab:eeg_imagenet}
\end{table*}

\subsection{Prototype-Based Decoding}

\paragraph{Retrieval.}
At inference, the EEG query and each gallery candidate are represented by their prototype-reconstructed embeddings. For an EEG trial $X$ and a candidate image $I_j$, the similarity score is
\begin{equation}
  s(X, I_j) = \cos(\zhate, \zhatv^j) = \frac{\zhate \cdot \zhatv^j}{\|\zhate\| \|\zhatv^j\|}.
\end{equation}
Gallery candidates are ranked by descending score, and Top-1 and Top-5 accuracy measure retrieval performance. Both the EEG query and visual gallery pass through the codebook, so ranking compares their learned prototype decompositions.

Gallery representations are computed once with the trained visual projection and codebook. Test-time processing then requires only EEG encoding, assignment, and similarity evaluation. The scoring function has no classifier tied to a fixed vocabulary, so the same interface supports the 200-way image gallery and the 80-way class gallery.

For the image gallery, each candidate is encoded with the same visual projection used during training and then reconstructed as $\zhatv$. In the class-level protocol, the gallery contains one representative for each class; in image-level retrieval, it contains the held-out candidate images. The similarity rule itself is unchanged.

\paragraph{Reconstruction.}
The prototype-reconstructed EEG embedding $\zhate$ conditions a frozen generative decoder. This decoder does not affect codebook training. Reconstruction measures the information retained by the prototype bottleneck with pixel-level and semantic metrics.

Retrieval and reconstruction use the same learned code. Retrieval tests candidate ordering; reconstruction tests which appearance and semantic information can be rendered. Neither task retrains the codebook. A difference between their outcomes therefore reflects what the fixed code preserves for each readout, rather than a separate representation learned for generation.

\section{Experiments}

\subsection{Datasets and Implementation Details}

\paragraph{Datasets.}
We evaluate on two EEG visual-decoding datasets. THINGS-EEG2~\citep{gifford2022things} contains 10 subjects, 16,740 natural images from 200 concepts, and 63-channel EEG recorded at 1000 Hz after 50 ms presentations. Each concept contains several image exemplars, and the model receives one EEG trial at a time. In \emph{intra-subject} evaluation, an 80/20 split trains and tests on one subject using disjoint exemplars of the same concepts. In \emph{inter-subject} evaluation, one subject is held out and the remaining nine provide training data. Both use a 200-way gallery with one held-out exemplar per concept. This design tests image retrieval rather than only predicting a concept label.

EEG-ImageNet~\citep{spampinato2017} contains 8 subjects viewing 40,000 images from 80 classes. Each image is presented for 500 ms, and EEG is recorded from 128 channels at 1000 Hz. We follow its 80-way class-retrieval protocol with disjoint training and test images and one representative per class. ProtoEEG is retrained rather than transferred zero-shot. This task retrieves classes rather than THINGS-EEG2 image instances, so scores across the datasets are not directly comparable.

\paragraph{Brain Encoders.}
The EEG encoder applies temporal convolution with a 25 ms kernel, spatial convolution over channels, and self-attention to each trial $X \in \R^{C \times T}$. A two-layer MLP $g_e$ projects the normalized 1024-dimensional output to $d=512$. The encoder contains no subject-specific parameters and receives no subject identifier. Differences between intra-subject and inter-subject results therefore arise without an explicit subject-adaptation module.

\paragraph{Visual Teachers.}
We use frozen CLIP ViT-L/14 and ResNet-50 teachers, which define two method families. The trainable head $g_v$ maps their features to the shared $d=512$ space. We compare methods only within a family so teacher capacity is controlled. Scores across the two families are reported separately because their frozen features have different capacity and geometry.

\paragraph{Training and Evaluation Details.}
The $K=64$ codebook is initialized from training-image embeddings. We train for 100 epochs with AdamW at a learning rate of $10^{-4}$ and a batch size of 128. We set $\tau=0.07$ and initialize the learned $\tau'$ to 0.07. The loss weights are $\lambda_{\mathrm{kl}}=0.5$, $\lambda_{\mathrm{inst}}=0.3$, and $\lambda_{\mathrm{div}}=0.1$. Results average five seeds before averaging subjects; standard deviations appear in the supplement. Table~\ref{tab:ablation} averages all ten THINGS-EEG2 subjects. Reconstruction uses a frozen SDXL IP-Adapter conditioned on $\zhate$.

\paragraph{Baselines.}
We compare with recent EEG visual decoding methods. NICE~\citep{song2024nice} aligns EEG with CLIP via noise-contrastive estimation. ATM~\citep{li2024atm} uses an adaptive teacher module with diffusion-based reconstruction. AVDE~\citep{dai2026avde} employs autoregressive visual decoding. NeuroBridge~\citep{zhang2026neurobridge} applies cognitive alignment for cross-subject transfer. All baselines use dense instance-level alignment without explicit prototype decomposition.

\subsection{Retrieval Performance}

\paragraph{Retrieval on THINGS-EEG2.}
Table~\ref{tab:retrieval} presents 200-way retrieval results for intra-subject evaluation, in which training and testing use one subject, and inter-subject evaluation, in which one subject is left out for test.

In intra-subject evaluation, ProtoEEG reaches 37.7/68.9 Top-1/Top-5 with CLIP, exceeding AVDE by +7.7/+10.7 percentage points, and 64.0/91.5 with RN50, exceeding NeuroBridge by +0.8/+1.6 pp. The smaller RN50 margin reflects its stronger baseline. Subject rows reveal complementary patterns: CLIP improves over AVDE for all ten subjects, whereas RN50 gains occur for six of ten, making the CLIP improvement more broadly distributed.

In inter-subject evaluation, ProtoEEG reaches 15.5/39.8 with CLIP, improving on AVDE by +1.3/+6.9 pp. With RN50, it reaches 22.0/48.4 and improves on NeuroBridge by +3.0/+2.5 pp. The fall from 64.0\% to 22.0\% RN50 Top-1 shows that cross-subject shift remains substantial, although the codebook preserves its within-family advantage without subject-specific calibration. For CLIP, the larger Top-5 gain suggests that the correct concept is retained in the shortlist more often than it is ranked first.

The per-subject rows qualify the averages. CLIP gains are consistent across participants, while the RN50 advantage is concentrated rather than uniform. Under inter-subject evaluation, both families retain a large spread across held-out subjects. Prototype alignment improves the target representation, but it does not remove differences in signal quality or neural response across people.

\paragraph{Retrieval on EEG-ImageNet.}
Table~\ref{tab:eeg_imagenet} reports the 80-way results after retraining ProtoEEG on EEG-ImageNet. ProtoEEG averages 44.8/73.8 Top-1/Top-5, exceeding AVDE by +20.4/+15.1 pp and winning on six of eight subjects. Top-1 ranges from 16.9\% to 77.1\%, exposing strong individual variation. Excluding the best subject still gives 40.2\% versus AVDE's 25.8\%, so one outlier does not explain the gain. Failures on Sub-02 and Sub-05 show that participant-specific difficulty persists.

\subsection{Image Reconstruction Quality}

\begin{figure*}[t]
  \centering
  \includegraphics[width=0.96\textwidth]{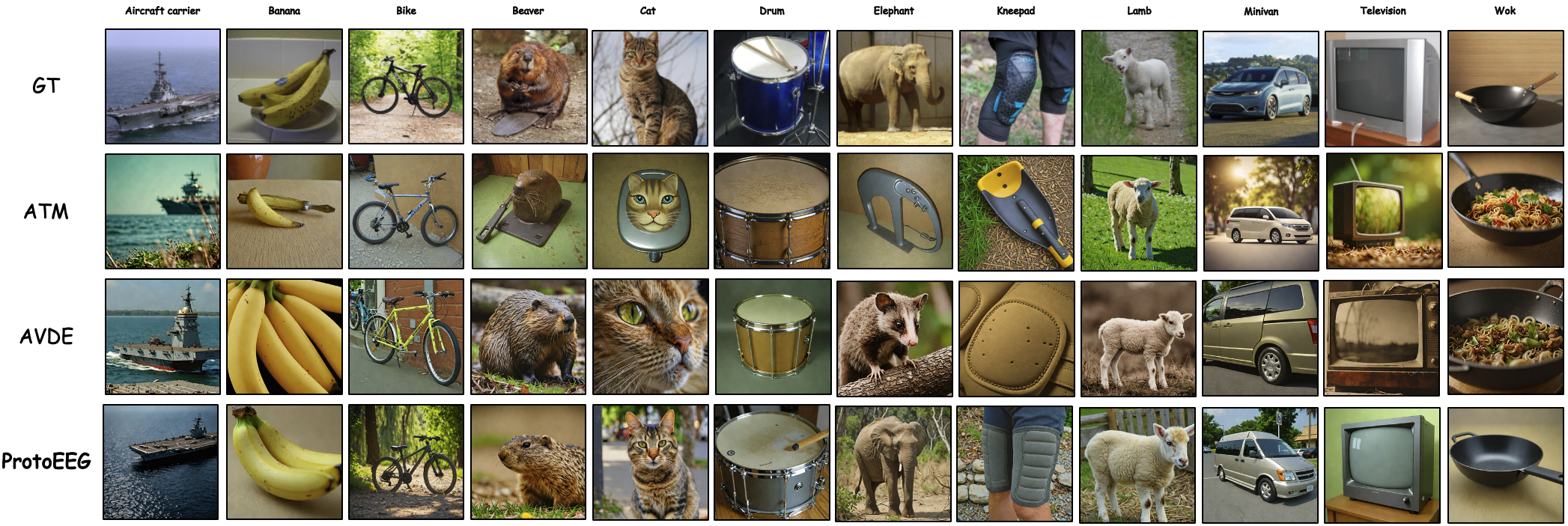}
  \caption{Qualitative reconstruction comparison on twelve THINGS-EEG2 test concepts.}
  \label{fig:gallery}
\end{figure*}

\begin{table}[t]
\centering
\scriptsize
\setlength{\tabcolsep}{0.7pt}
\renewcommand{\arraystretch}{1.0}
\begin{tabular*}{\columnwidth}{@{\extracolsep{\fill}}l*{7}{c}@{}}
\toprule
\textbf{Method} & \multicolumn{3}{c}{\textbf{Low-level}} & \multicolumn{4}{c}{\textbf{High-level}} \\
\cmidrule(lr){2-4}\cmidrule(lr){5-8}
& \textbf{PixCorr$\uparrow$} & \textbf{SSIM$\uparrow$} & \textbf{AlexNet(2)$\uparrow$} & \textbf{AlexNet(5)$\uparrow$} & \textbf{Inception$\uparrow$} & \textbf{CLIP$\uparrow$} & \textbf{SwAV$\downarrow$} \\
\midrule
ATM & 0.130 & 0.339 & 0.740 & 0.822 & 0.685 & 0.716 & 0.599 \\
AVDE & 0.147 & 0.366 & 0.766 & 0.835 & 0.724 & 0.747 & 0.586 \\
\textbf{ProtoEEG} & \textbf{0.154} & \textbf{0.390} & \textbf{0.779} & \textbf{0.855} & \textbf{0.743} & \textbf{0.768} & \textbf{0.506} \\
\bottomrule
\end{tabular*}
\caption{Reconstruction performance on THINGS-EEG2. ProtoEEG reports five-seed means over ten subjects.}
\label{tab:recon}
\end{table}

Table~\ref{tab:recon} separates low-level structure from high-level semantic identification. ProtoEEG leads all seven evaluators: over AVDE, it gains +0.007 PixCorr, +0.024 SSIM, +0.013 AlexNet(2), +0.020 AlexNet(5), +0.019 Inception, and +0.021 CLIP, while reducing SwAV distance by 0.080. The agreement across metric families shows that the bottleneck retains both appearance and category information. Figure~\ref{fig:gallery} demonstrates that ProtoEEG reconstructions preserve semantic content and pixel-level details better than baselines. Dominant object categories are correctly identified, and visual attributes such as color, texture, and coarse shape are more faithfully reproduced compared to ATM and AVDE.

\subsection{Analysis of Prototype Mechanism}

\begin{figure}[t]
  \centering
  \includegraphics[width=0.94\columnwidth]{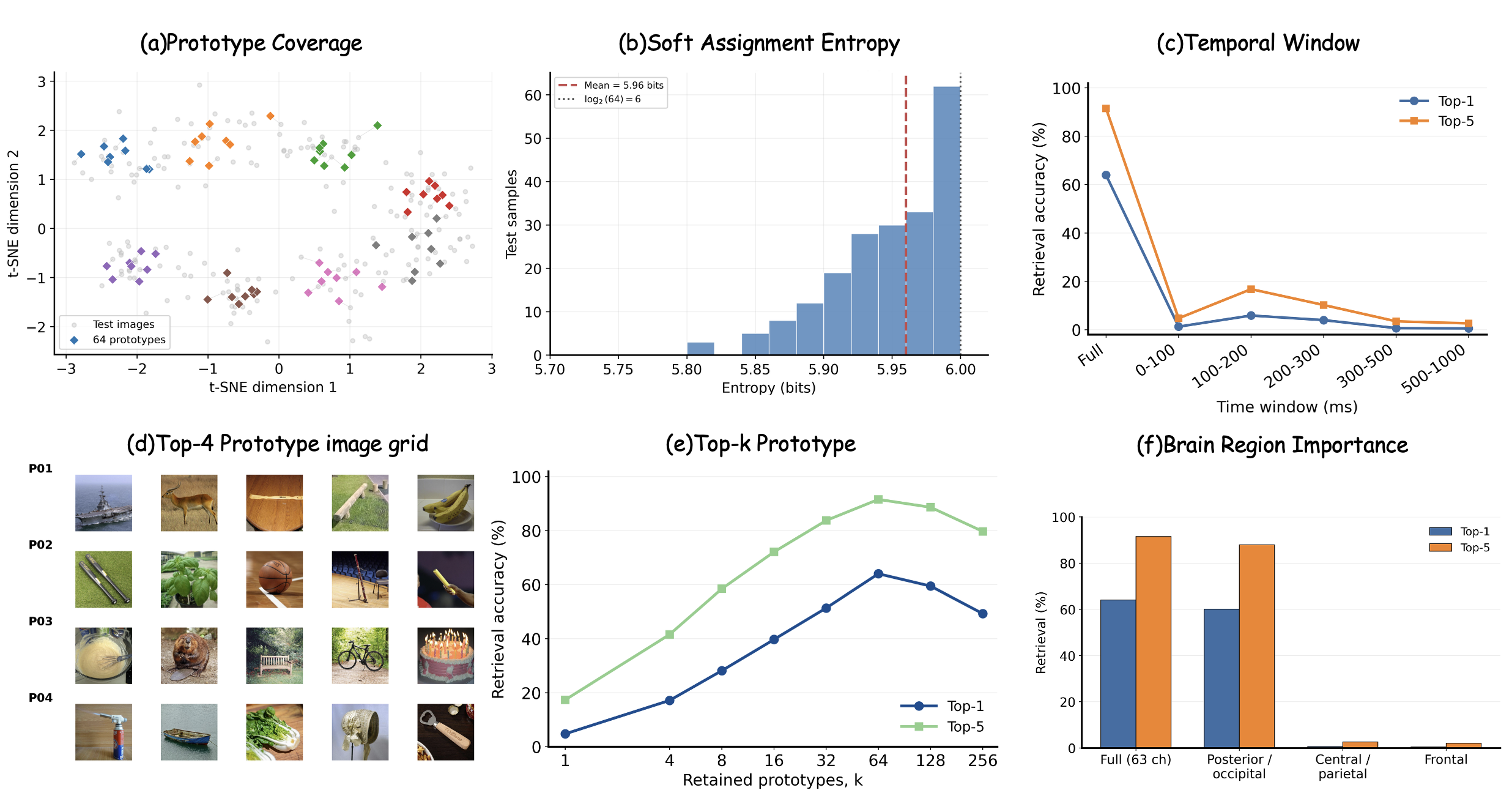}
  \caption{Prototype diagnostics over five seeds and ten subjects. Panels a--f show prototype coverage, assignment entropy, temporal windows, activating images, top-$k$ retention, and channel groups.}
  \label{fig:prototype_explanation}
\end{figure}

\begin{table}[t]
\centering
\setlength{\tabcolsep}{4pt}
\small
\begin{tabular*}{\columnwidth}{@{\extracolsep{\fill}}lccc@{}}
\toprule
\textbf{Configuration} & \textbf{Top-1} & \textbf{Top-5} & \textbf{$\Delta$T1} \\
\midrule
\multicolumn{4}{l}{\textit{Loss ablation}} \\
Full Loss & 64.0 & 91.5 & --- \\
w/o $\Lnce$ & 7.1  & 17.7 & $-56.9$ \\
w/o $\Lkl$  & 33.7 & 58.5 & $-30.3$ \\
w/o $\Linst$& 45.5 & 65.4 & $-18.5$ \\
w/o $\Ldiv$ & 51.6 & 77.1 & $-12.4$ \\
\midrule
\multicolumn{4}{l}{\textit{Assignment perturbation}} \\
Full prototypes, $K=64$ & 64.0 & 91.5 & --- \\
Raw $\ze$, no prototypes & 29.4 & 49.0 & $-34.6$ \\
Top-$k$, $k=48$ & 34.5 & 60.3 & $-29.5$ \\
Top-$k$, $k=16$ & 3.7  & 7.3  & $-60.3$ \\
Random-$k$, $k=16$ & 1.8  & 4.5  & $-62.2$ \\
Shuffled $\alpha$, $k=48$ & 0.5  & 2.0  & $-63.5$ \\
\bottomrule
\end{tabular*}
\caption{ProtoEEG diagnostics: five-seed means over all ten THINGS-EEG2 subjects.}
\label{tab:ablation}
\end{table}

Table~\ref{tab:ablation} presents loss ablations and assignment perturbations, averaged over all ten THINGS-EEG2 subjects under the intra-subject protocol.

\paragraph{Dominant Role of Prototype Contrastive Loss.}
Removing $\Lnce$ causes the largest Top-1 drop of $-56.9$ pp. Removing $\Lkl$, $\Linst$, and $\Ldiv$ reduces Top-1 by $-30.3$, $-18.5$, and $-12.4$ pp, respectively. Prototype contrast is the primary signal, while the remaining terms regularize it.

\paragraph{Necessity of Learned Prototype Structure.}
Bypassing prototypes drops Top-1 from 64.0\% to 29.4\%. Retaining the top 48 or 16 assignments gives 34.5\% or 3.7\%; selecting 16 at random gives 1.8\%. Shuffling weights across prototype identities reduces accuracy to 0.5\%, despite preserving their magnitudes. Retrieval therefore depends on both distributed weights and learned prototype identities.

\paragraph{Prototype Coverage and Assignment Entropy.}
Panel (a) of Figure~\ref{fig:prototype_explanation} shows coverage across several t-SNE regions. The learned prototypes distribute across the visual space rather than clustering in a small region. Mean assignment entropy is 5.96 bits, near the six-bit maximum for a 64-prototype codebook. The assignments are distributed rather than winner-take-all, meaning each EEG trial activates multiple prototypes with graded weights. The activating images for each prototype show semantic consistency within clusters but vary in viewpoint and background, characterizing prototypes as reusable visual factors rather than one-to-one class representatives.

\paragraph{Top-$k$ Retention and Temporal/Spatial EEG Evidence.}
For a separate $K=256$ checkpoint, we test which subset of assignments is necessary. Retention peaks at $k=64$ and declines at $k=128$ and 256. This suggests that increasing codebook size beyond 64 does not improve retrieval when all assignments are used, but retaining the top 64 from a larger set performs comparably to using all 64 from a smaller codebook. Panel (c) shows that the 100--200 ms post-stimulus window provides the strongest temporal evidence, consistent with visual processing latencies. Panel (f) shows that posterior and occipital channels contribute most strongly, reflecting their proximity to visual cortex.

\paragraph{Interpreting Assignment Structure.}
High entropy and top-$k$ retention describe different properties. The $K=64$ assignments avoid one-prototype collapse, showing that the model uses the full codebook capacity. The $K=256$ intervention shows that not every component is equally necessary: 64 prototypes capture most of the discriminative structure. Shuffling reduces performance to near chance, demonstrating that accuracy depends on which prototype receives each weight, not only on weight magnitude. The learned prototype identities therefore encode visual structure that the assignment mechanism exploits.

\FloatBarrier

\section{Conclusion}

We presented ProtoEEG, a framework that reformulates dense EEG--image alignment as soft composition over a shared visual codebook. A trainable EEG encoder and frozen visual teacher produce assignment profiles over the same prototypes, and the reconstructed EEG representation supports retrieval and conditions a fixed image generator. Prototype contrast provides the main cross-modal signal, while distribution alignment, instance anchoring, and diversity regularization coordinate codebook use and prevent collapse. Across five seeds, ProtoEEG reaches 64.0\% Top-1 and 91.5\% Top-5 on THINGS-EEG2, obtains 44.8\% and 73.8\% after retraining on EEG-ImageNet, and leads seven reconstruction metrics. Retention and shuffling experiments further show that decoding depends on distributed weights and learned prototype identities. ProtoEEG makes target structure an explicit design choice for EEG visual decoding. Future work should improve cross-subject calibration and retain finer instance details without losing the inspectable prototype interface.

\bibliography{references}

@inproceedings{bai2024dreamdiffusion,
  title={DreamDiffusion: High-quality EEG-to-image generation with temporal masked signal modeling and CLIP alignment},
  author={Bai, Yunpeng and Wang, Xintao and Cao, Yan-Pei and Ge, Yixiao and Yuan, Chun and Shan, Ying},
  booktitle={European Conference on Computer Vision},
  pages={472--488},
  year={2024},
  organization={Springer}
}

@inproceedings{benchetrit2024braindecoding,
  title={Brain decoding: toward real-time reconstruction of visual perception},
  author={Benchetrit, Yohann and Banville, Hubert and King, Jean-R{\'e}mi},
  booktitle={International Conference on Learning Representations},
  volume={2024},
  pages={7846--7858},
  year={2024}
}

@inproceedings{caron2018deepcluster,
  title={Deep clustering for unsupervised learning of visual features},
  author={Caron, Mathilde and Bojanowski, Piotr and Joulin, Armand and Douze, Matthijs},
  booktitle={Proceedings of the European conference on computer vision (ECCV)},
  pages={132--149},
  year={2018}
}

@inproceedings{caron2020swav,
  author    = {Mathilde Caron and Ishan Misra and Julien Mairal and
               Priya Goyal and Piotr Bojanowski and Armand Joulin},
  title     = {Unsupervised Learning of Visual Features by Contrasting
               Cluster Assignments},
  booktitle = {Advances in Neural Information Processing Systems (NeurIPS)},
  year      = {2020},
}

@inproceedings{chen2019protopnet,
  author    = {Chaofan Chen and Oscar Li and Chaofan Tao and
               Alina Jade Barnett and Jonathan Su and Cynthia Rudin},
  title     = {This Looks Like That: Deep Learning for Interpretable
               Image Recognition},
  booktitle = {Advances in Neural Information Processing Systems (NeurIPS)},
  year      = {2019},
  pages     = {8930--8941},
}

@inproceedings{cheng2023prior,
  title={Prior: Prototype representation joint learning from medical images and reports},
  author={Cheng, Pujin and Lin, Li and Lyu, Junyan and Huang, Yijin and Luo, Wenhan and Tang, Xiaoying},
  booktitle={Proceedings of the IEEE/CVF international conference on computer vision},
  pages={21361--21371},
  year={2023}
}

@inproceedings{dai2026avde,
  author    = {Sicheng Dai and Hongwang Xiao and Shan Yu and Qiwei Ye},
  title     = {Autoregressive Visual Decoding from {EEG} Signals},
  booktitle = {International Conference on Learning Representations (ICLR)},
  year      = {2026},
}

@inproceedings{fu2025brainvis,
  title={Brainvis: Exploring the bridge between brain and visual signals via image reconstruction},
  author={Fu, Honghao and Wang, Hao and Chin, Jing Jih and Shen, Zhiqi},
  booktitle={ICASSP 2025-2025 IEEE International Conference on Acoustics, Speech and Signal Processing (ICASSP)},
  pages={1--5},
  year={2025},
  organization={IEEE}
}

@article{gifford2022things,
  title={A large and rich EEG dataset for modeling human visual object recognition},
  author={Gifford, Alessandro T and Dwivedi, Kshitij and Roig, Gemma and Cichy, Radoslaw M},
  journal={NeuroImage},
  volume={264},
  pages={119754},
  year={2022},
  publisher={Elsevier}
}

@inproceedings{li2021pcl,
  title={Prototypical contrastive learning of unsupervised representations},
  author={Li, Junnan and Zhou, Pan and Xiong, Caiming and Hoi, Steven CH},
  journal={arXiv preprint arXiv:2005.04966},
  year={2020}
}

@inproceedings{li2024atm,
  title={Visual decoding and reconstruction via eeg embeddings with guided diffusion},
  author={Li, Dongyang and Wei, Chen and Li, Shiying and Zou, Jiachen and Qin, Haoyang and Liu, Quanying},
  journal={arXiv preprint arXiv:2403.07721},
  year={2024}
}

@inproceedings{li2025brainflora,
  author    = {Dongyang Li and Haoyang Qin and Mingyang Wu and
               Chen Wei and Quanying Liu},
  title     = {{BrainFLORA}: Uncovering Brain Concept Representation via
               Multimodal Neural Embeddings},
  booktitle = {Proceedings of the 33rd ACM International Conference on
               Multimedia (ACM MM)},
  year      = {2025},
}

@inproceedings{liang2022mindthegap,
  title={Mind the gap: Understanding the modality gap in multi-modal contrastive representation learning},
  author={Liang, Weixin and Zhang, Yuhui and Kwon, Yongchan and Yeung, Serena and Zou, James},
  journal={arXiv preprint arXiv:2203.02053},
  year={2022}
}

@article{liu2025vieeg,
  title={ViEEG: Hierarchical Visual Neural Representation for EEG Brain Decoding},
  author={Liu, Minxu and Guan, Donghai and Zheng, Chuhang and Tian, Chunwei and Wen, Jie and Zhu, Qi},
  journal={arXiv preprint arXiv:2505.12408},
  year={2025}
}

@article{oord2018infonce,
  title={Representation learning with contrastive predictive coding},
  author={Oord, Aaron van den and Li, Yazhe and Vinyals, Oriol},
  journal={arXiv preprint arXiv:1807.03748},
  year={2018}
}

@article{ozcelik2023braindiffuser,
  title={Natural scene reconstruction from fMRI signals using generative latent diffusion},
  author={Ozcelik, Furkan and VanRullen, Rufin},
  journal={Scientific Reports},
  volume={13},
  number={1},
  pages={15666},
  year={2023},
  publisher={Nature Publishing Group UK London}
}

@article{palazzo2020,
  title={Decoding brain representations by multimodal learning of neural activity and visual features},
  author={Palazzo, Simone and Spampinato, Concetto and Kavasidis, Isaak and Giordano, Daniela and Schmidt, Joseph and Shah, Mubarak},
  journal={IEEE Transactions on Pattern Analysis and Machine Intelligence},
  volume={43},
  number={11},
  pages={3833--3849},
  year={2020},
  publisher={IEEE}
}

@article{qin2023capro,
  title={Capro: webly supervised learning with cross-modality aligned prototypes},
  author={Qin, Yulei and Chen, Xingyu and Shen, Yunhang and Fu, Chaoyou and Gu, Yun and Li, Ke and Sun, Xing and Ji, Rongrong},
  journal={Advances in Neural Information Processing Systems},
  volume={36},
  pages={53565--53592},
  year={2023}
}

@inproceedings{rajabi2025humanaligned,
  title={Human-aligned image models improve visual decoding from the brain},
  author={Rajabi, Nona and Ribeiro, Ant{\^o}nio H and Vasco, Miguel and Taleb, Farzaneh and Bj{\"o}rkman, M{\aa}rten and Kragic, Danica},
  booktitle={Forty-second International Conference on Machine Learning},
  year={2025}
}

@inproceedings{scotti2023mindeye,
  title={Reconstructing the mind's eye: fmri-to-image with contrastive learning and diffusion priors},
  author={Scotti, Paul and Banerjee, Atmadeep and Goode, Jimmie and Shabalin, Stepan and Nguyen, Alex and Dempster, Aidan and Verlinde, Nathalie and Yundler, Elad and Weisberg, David and Norman, Kenneth and others},
  journal={Advances in Neural Information Processing Systems},
  volume={36},
  pages={24705--24728},
  year={2023}
}

@article{scotti2024mindeye2,
  title={Mindeye2: Shared-subject models enable fmri-to-image with 1 hour of data},
  author={Scotti, Paul S and Tripathy, Mihir and Villanueva, Cesar Kadir Torrico and Kneeland, Reese and Chen, Tong and Narang, Ashutosh and Santhirasegaran, Charan and Xu, Jonathan and Naselaris, Thomas and Norman, Kenneth A and others},
  journal={arXiv preprint arXiv:2403.11207},
  year={2024}
}

@inproceedings{singh2024robust,
  title={Learning robust deep visual representations from EEG brain recordings},
  author={Singh, Prajwal and Dalal, Dwip and Vashishtha, Gautam and Miyapuram, Krishna and Raman, Shanmuganathan},
  booktitle={Proceedings of the IEEE/CVF Winter Conference on Applications of Computer Vision},
  pages={7553--7562},
  year={2024}
}

@article{snell2017prototypical,
  title={Prototypical networks for few-shot learning},
  author={Snell, Jake and Swersky, Kevin and Zemel, Richard},
  journal={Advances in neural information processing systems},
  volume={30},
  year={2017}
}

@inproceedings{song2024nice,
  title={Decoding natural images from eeg for object recognition},
  author={Song, Yonghao and Liu, Bingchuan and Li, Xiang and Shi, Nanlin and Wang, Yijun and Gao, Xiaorong},
  journal={arXiv preprint arXiv:2308.13234},
  year={2023}
}

@inproceedings{spampinato2017,
  title={Deep learning human mind for automated visual classification},
  author={Spampinato, Concetto and Palazzo, Simone and Kavasidis, Isaak and Giordano, Daniela and Souly, Nasim and Shah, Mubarak},
  booktitle={Proceedings of the IEEE Conference on Computer Vision and Pattern Recognition},
  pages={6809--6817},
  year={2017}
}

@inproceedings{wu2025ubp,
  title={Bridging the vision-brain gap with an uncertainty-aware blur prior},
  author={Wu, Haitao and Li, Qing and Zhang, Changqing and He, Zhen and Ying, Xiaomin},
  booktitle={Proceedings of the IEEE/CVF Conference on Computer Vision and Pattern Recognition},
  pages={2246--2257},
  year={2025}
}

@inproceedings{wu2026ats,
  title={Shrinking the teacher: An adaptive teaching paradigm for asymmetric eeg-vision alignment},
  author={Wu, Lukun and Li, Jie and Ren, Ziqi and Zhang, Kaifan and Gao, Xinbo},
  booktitle={Proceedings of the AAAI Conference on Artificial Intelligence},
  volume={40},
  number={21},
  pages={17859--17867},
  year={2026}
}

@inproceedings{zhang2025cognitioncapturer,
  title={Cognitioncapturer: Decoding visual stimuli from human eeg signal with multimodal information},
  author={Zhang, Kaifan and He, Lihuo and Jiang, Xin and Lu, Wen and Wang, Di and Gao, Xinbo},
  booktitle={Proceedings of the AAAI Conference on Artificial Intelligence},
  volume={39},
  number={13},
  pages={14486--14493},
  year={2025}
}

@inproceedings{zhang2026neurobridge,
  title={Neurobridge: Bio-inspired self-supervised eeg-to-image decoding via cognitive priors and bidirectional semantic alignment},
  author={Zhang, Wenjiang and Wang, Sifeng and Su, Yuwei and Li, Xinyu and Zhang, Chen and Zhong, Suyu},
  booktitle={Proceedings of the AAAI Conference on Artificial Intelligence},
  volume={40},
  number={21},
  pages={18028--18036},
  year={2026}
}

@inproceedings{yupatchcode,
  title={PATCHCODE: Discrete Latent Predictive Learning for EEG Foundation Model},
  author={Yu, Kieren and Liu, Ziyang and Huang, Chang and Wu, Kaishun},
  booktitle={Forty-third International Conference on Machine Learning},
  year={2026}
}

@inproceedings{jiang2024large,
  title={Large brain model for learning generic representations with tremendous EEG data in BCI},
  author={Jiang, Wei-Bang and Zhao, Liming and Lu, Bao-Liang},
  booktitle={International Conference on Learning Representations},
  volume={2024},
  pages={16405--16426},
  year={2024}
}

@article{zhou2026csbrain,
  title={Csbrain: A cross-scale spatiotemporal brain foundation model for eeg decoding},
  author={Zhou, Yuchen and Wu, Jiamin and Ren, Zichen and Yao, Zhouheng and Lu, Weiheng and Peng, Kunyu and Zheng, Qihao and Song, Chunfeng and Ouyang, Wanli and Gou, Chao},
  journal={Advances in Neural Information Processing Systems},
  volume={38},
  pages={87150--87195},
  year={2026}
}

@inproceedings{wang2025cbramod,
  title={Cbramod: A criss-cross brain foundation model for eeg decoding},
  author={Wang, Jiquan and Zhao, Sha and Luo, Zhiling and Zhou, Yangxuan and Jiang, Haiteng and Li, Shijian and Li, Tao and Pan, Gang},
  booktitle={International conference on learning representations},
  volume={2025},
  pages={75310--75346},
  year={2025}
}

@article{el2026reve,
  title={REVE: A foundation model for EEG-adapting to any setup with large-scale pretraining on 25,000 subjects},
  author={El Ouahidi, Yassine and Lys, Jonathan and Th{\"o}lke, Philipp and Farrugia, Nicolas and Pasdeloup, Bastien and Gripon, Vincent and Jerbi, Karim and Lioi, Giulia},
  journal={Advances in Neural Information Processing Systems},
  volume={38},
  pages={22541--22577},
  year={2026}
}

@inproceedings{liu2024itransformer,
  author    = {Yong Liu and Tengge Hu and Haoran Zhang and Haixu Wu and
               Shiyu Wang and Lintao Ma and Mingsheng Long},
  title     = {{iTransformer}: Inverted Transformers Are Effective for
               Time Series Forecasting},
  booktitle = {International Conference on Learning Representations (ICLR)},
  year      = {2024},
}

@article{podell2024sdxl,
  author    = {Dustin Podell and Zion English and Kyle Lacey and
               Andreas Blattmann and Tim Dockhorn and Jonas M{\"u}ller and
               Joe Penna and Robin Rombach},
  title     = {{SDXL}: Improving Latent Diffusion Models for
               High-Resolution Image Synthesis},
  journal   = {arXiv preprint arXiv:2307.01952},
  year      = {2023},
}

@inproceedings{radford2021clip,
  author    = {Alec Radford and Jong Wook Kim and Chris Hallacy and
               Aditya Ramesh and Gabriel Goh and Sandhini Agarwal and
               Girish Sastry and Amanda Askell and Pamela Mishkin and
               Jack Clark and Gretchen Krueger and Ilya Sutskever},
  title     = {Learning Transferable Visual Models From Natural Language
               Supervision},
  booktitle = {Proceedings of the 38th International Conference on
               Machine Learning (ICML)},
  year      = {2021},
  pages     = {8748--8763},
}

@article{ye2023ipadapter,
  author    = {Hu Ye and Jun Zhang and Sibo Liu and Xiao Han and Wei Yang},
  title     = {{IP-Adapter}: Text Compatible Image Prompt Adapter for
               Text-to-Image Diffusion Models},
  journal   = {arXiv preprint arXiv:2308.06721},
  year      = {2023},
}

\clearpage
\appendix
\setcounter{secnumdepth}{2}
\setcounter{table}{0}
\renewcommand{\thetable}{S\arabic{table}}
\setlength{\tabcolsep}{4pt}
\section*{Supplementary Material}
\section*{Overview}

This supplementary document provides the full experimental and evaluation protocol, subject-wise reconstruction results for ATM, AVDE, and ProtoEEG on THINGS-EEG2 in Table~\ref{tab:subject_reconstruction}, the mapping between the archived ProtoEEG subject export and the matched main-paper aggregate, and the efficiency measurements together with their scope.

\section{Experimental Setup and Evaluation Protocol}
\label{app:setup}

\subsection{Datasets and evaluation splits}
THINGS-EEG2~\citep{gifford2022things} contains EEG from 10 participants viewing images drawn from 1,654 object concepts. We use the standard held-out set for 200-way retrieval, with one gallery image per test concept. Subject-dependent (SD) models are trained and evaluated within subject; subject-independent (SI) evaluation follows leave-one-subject-out (LOSO) training. EEG-ImageNet~\citep{spampinato2017} contains 8 subjects and 80 ImageNet classes. Its primary measure is class-averaged Top-1/Top-5 accuracy; trial-level assignment is reported separately and is not compared with class-level baseline scores.

\subsection{Random seeds and aggregation}
All ProtoEEG quantitative results in the main paper are arithmetic means over five independently initialised runs. We first average the five runs separately for each subject and metric; Avg/Mean values then average the underlying subject means before display rounding. Tables~\ref{tab:retrieval} and~\ref{tab:eeg_imagenet} report the displayed subject rows, while Table~\ref{tab:ablation} and the quantitative panels of Figure~\ref{fig:prototype_explanation} additionally average all ten THINGS-EEG2 subjects. No best-seed selection is used.

\subsection{Model configuration and training}
The EEG encoder is an iTransformer~\citep{liu2024itransformer} operating on $63\times250$ input tensors and projecting to a $d=1024$ shared space through a two-layer MLP. ProtoEEG-CLIP uses a frozen CLIP ViT-L/14 visual teacher~\citep{radford2021clip}; ProtoEEG-RN50 uses a frozen compressed RN50 teacher matching the bandwidth-adapted baseline family. The shared codebook contains $K=64$ L2-normalised prototypes with temperature $\tau=0.07$.

We train with AdamW ($\beta_1=0.9$, $\beta_2=0.999$, weight decay $10^{-4}$), learning rate $3\times10^{-4}$ with cosine decay and 20-epoch warm-up, batch size 1024, and 300 epochs. Loss weights are $\lambda_{\mathrm{kl}}=0.5$, $\lambda_{\mathrm{inst}}=0.3$, and $\lambda_{\mathrm{div}}=0.1$. Training uses four A100 80\,GB GPUs, FP16 forward passes, FP32 gradient accumulation, Gaussian-noise augmentation with $\sigma=0.1$, and channel dropout of 0.05.

\subsection{Baselines, metrics, and fairness}
Comparisons are separated by visual teacher. The CLIP family contains NICE~\citep{song2024nice}, ATM~\citep{li2024atm}, and AVDE~\citep{dai2026avde}; the RN50/compressed-teacher family contains UBP~\citep{wu2025ubp}, ATS~\citep{wu2026ats}, and NeuroBridge~\citep{zhang2026neurobridge}. All claimed improvements are within family.

Retrieval reports Top-1 and Top-5 accuracy in the fixed 200-image gallery. Reconstruction reports Pixel Correlation and SSIM for low-level fidelity; CLIP, AlexNet-2, AlexNet-5, and Inception identification for semantic correspondence; and SwAV distance, for which lower is better. The prototype-reconstructed embedding conditions a frozen SDXL IP-Adapter~\citep{ye2023ipadapter,podell2024sdxl} using 50 DDIM steps, guidance 7.5, an empty text prompt, and evaluation seed 42.

\subsection{Gallery, LOSO, and diagnostic protocols}
Each THINGS-EEG2 gallery image is first mapped to $z_v$ by the frozen visual backbone and trainable projection head, assigned through the shared codebook as $\alpha_v=\operatorname{softmax}(z_vP^\top/\tau)$, and reconstructed as $\hat{z}_v=\alpha_vP$. The prototype-reconstructed EEG query $\hat{z}_e=\alpha_eP$ is ranked against these $\hat{z}_v$ gallery representations by cosine similarity. In LOSO evaluation, the model is trained on nine subjects and evaluated on the remaining subject without fine-tuning or subject-specific test-time adaptation. Loss ablations, assignment perturbations, temporal-window masking, channel-group masking, and the quantitative prototype diagnostics follow the same five-seed protocol and are averaged over all ten THINGS-EEG2 subjects. Figure~\ref{fig:prototype_explanation}(e) alone uses a separate $K=256$ checkpoint family, as stated in the main text.

\clearpage
\twocolumn[
\begin{minipage}{\textwidth}
\section{Reconstruction Details on THINGS-EEG2}
\label{app:subject_reconstruction}

\noindent Table~\ref{tab:subject_reconstruction} preserves all available subject-wise reconstruction results. The ATM and AVDE rows are paper-reported values from the AVDE supplementary material~\citep{dai2026avde}, while the ProtoEEG subject rows reproduce the archived per-subject export. Because that export used CLIPScore and SwAV identification, panel (c) separately reproduces the matched five-seed, ten-subject ProtoEEG aggregate from main-paper Table~\ref{tab:recon} under CLIP identification and SwAV distance. Panel (c), rather than an arithmetic combination across incompatible evaluator definitions, is the reference for main-paper comparisons.

\begin{center}
\centering
\small
\setlength{\tabcolsep}{5pt}
\renewcommand{\arraystretch}{1.0}

\noindent\textbf{(a) ATM and AVDE: paper-reported evaluation}\par\vspace{-2pt}
\begin{tabular*}{\textwidth}{@{\extracolsep{\fill}}llccccccc@{}}
\toprule
\textbf{Method} & \textbf{Subject} &
\textbf{PixCorr} $\uparrow$ & \textbf{SSIM} $\uparrow$ &
\textbf{AlexNet(2)} $\uparrow$ & \textbf{AlexNet(5)} $\uparrow$ &
\textbf{Inception} $\uparrow$ & \textbf{CLIP-ID} $\uparrow$ &
\textbf{SwAV} $\downarrow$ \\
\midrule
\multirow{11}{*}{ATM}
& Sub-01 & 0.136 & 0.354 & 0.720 & 0.831 & 0.710 & 0.728 & 0.589 \\
& Sub-02 & 0.108 & 0.301 & 0.688 & 0.775 & 0.657 & 0.705 & 0.618 \\
& Sub-03 & 0.145 & 0.340 & 0.747 & 0.858 & 0.751 & 0.759 & 0.590 \\
& Sub-04 & 0.075 & 0.305 & 0.729 & 0.782 & 0.694 & 0.688 & 0.623 \\
& Sub-05 & 0.133 & 0.352 & 0.738 & 0.791 & 0.637 & 0.707 & 0.612 \\
& Sub-06 & 0.140 & 0.363 & 0.753 & 0.834 & 0.686 & 0.742 & 0.595 \\
& Sub-07 & 0.155 & 0.360 & 0.769 & 0.851 & 0.684 & 0.697 & 0.608 \\
& Sub-08 & 0.163 & 0.345 & 0.786 & 0.868 & 0.730 & 0.770 & 0.575 \\
& Sub-09 & 0.134 & 0.338 & 0.736 & 0.788 & 0.626 & 0.652 & 0.606 \\
& Sub-10 & 0.112 & 0.329 & 0.735 & 0.843 & 0.674 & 0.708 & 0.570 \\
& \textbf{Mean} & \textbf{0.130} & \textbf{0.339} & \textbf{0.740} & \textbf{0.822} & \textbf{0.685} & \textbf{0.716} & \textbf{0.599} \\
\midrule
\multirow{11}{*}{AVDE}
& Sub-01 & 0.138 & 0.362 & 0.750 & 0.837 & 0.725 & 0.721 & 0.609 \\
& Sub-02 & 0.120 & 0.388 & 0.754 & 0.811 & 0.710 & 0.737 & 0.604 \\
& Sub-03 & 0.168 & 0.379 & 0.763 & 0.827 & 0.704 & 0.746 & 0.603 \\
& Sub-04 & 0.126 & 0.361 & 0.748 & 0.835 & 0.731 & 0.765 & 0.593 \\
& Sub-05 & 0.145 & 0.373 & 0.743 & 0.814 & 0.691 & 0.732 & 0.586 \\
& Sub-06 & 0.128 & 0.364 & 0.768 & 0.832 & 0.707 & 0.717 & 0.584 \\
& Sub-07 & 0.126 & 0.351 & 0.760 & 0.827 & 0.733 & 0.731 & 0.588 \\
& Sub-08 & 0.188 & 0.396 & 0.817 & 0.889 & 0.765 & 0.795 & 0.557 \\
& Sub-09 & 0.156 & 0.339 & 0.753 & 0.810 & 0.713 & 0.726 & 0.600 \\
& Sub-10 & 0.173 & 0.349 & 0.807 & 0.871 & 0.764 & 0.800 & 0.543 \\
& \textbf{Mean} & \textbf{0.147} & \textbf{0.366} & \textbf{0.766} & \textbf{0.835} & \textbf{0.724} & \textbf{0.747} & \textbf{0.586} \\
\bottomrule
\end{tabular*}

\vspace{4pt}
\noindent\textbf{(b) ProtoEEG: archived subject-wise export}\par\vspace{-2pt}
\begin{tabular*}{\textwidth}{@{\extracolsep{\fill}}lccccccc@{}}
\toprule
\textbf{Subject} &
\textbf{PixCorr} $\uparrow$ & \textbf{SSIM} $\uparrow$ &
\textbf{AlexNet(2)} $\uparrow$ & \textbf{AlexNet(5)} $\uparrow$ &
\textbf{Inception} $\uparrow$ & \textbf{CLIPScore} $\uparrow$ &
\textbf{SwAV-ID} $\uparrow$ \\
\midrule
Sub-01 & 0.149 & 0.255 & 0.769 & 0.848 & 0.758 & 23.62 & 0.826 \\
Sub-02 & 0.140 & 0.265 & 0.782 & 0.858 & 0.749 & 22.70 & 0.829 \\
Sub-03 & 0.157 & 0.246 & 0.777 & 0.867 & 0.742 & 22.09 & 0.823 \\
Sub-04 & 0.137 & 0.261 & 0.779 & 0.857 & 0.731 & 22.65 & 0.842 \\
Sub-05 & 0.122 & 0.252 & 0.749 & 0.803 & 0.684 & 19.48 & 0.759 \\
Sub-06 & 0.148 & 0.254 & 0.773 & 0.846 & 0.700 & 21.95 & 0.803 \\
Sub-07 & 0.134 & 0.267 & 0.795 & 0.857 & 0.763 & 22.46 & 0.836 \\
Sub-08 & 0.145 & 0.259 & 0.819 & 0.897 & 0.764 & 23.41 & 0.864 \\
Sub-09 & 0.149 & 0.269 & 0.743 & 0.827 & 0.706 & 20.03 & 0.804 \\
Sub-10 & 0.156 & 0.273 & 0.806 & 0.894 & 0.832 & 25.39 & 0.892 \\
\textbf{Mean} & \textbf{0.144} & \textbf{0.260} & \textbf{0.779} & \textbf{0.855} & \textbf{0.743} & \textbf{22.38} & \textbf{0.828} \\
\bottomrule
\end{tabular*}

\vspace{4pt}
\noindent\textbf{(c) ProtoEEG: matched main-paper aggregate}\par\vspace{-2pt}
\begin{tabular*}{\textwidth}{@{\extracolsep{\fill}}lccccccc@{}}
\toprule
\textbf{Subject} &
\textbf{PixCorr} $\uparrow$ & \textbf{SSIM} $\uparrow$ &
\textbf{AlexNet(2)} $\uparrow$ & \textbf{AlexNet(5)} $\uparrow$ &
\textbf{Inception} $\uparrow$ & \textbf{CLIP} $\uparrow$ &
\textbf{SwAV} $\downarrow$ \\
\midrule
\textbf{Mean} & \textbf{0.154} & \textbf{0.390} & \textbf{0.779} & \textbf{0.855} & \textbf{0.743} & \textbf{0.768} & \textbf{0.506} \\
\bottomrule
\end{tabular*}
\captionof{table}{Subject-wise reconstruction details on THINGS-EEG2. Panel (a) reports ATM and AVDE under their paper-reported evaluators. Panel (b) preserves the archived ProtoEEG subject export, whose final two columns are CLIPScore and SwAV identification. Panel (c) gives the five-seed, ten-subject ProtoEEG aggregate used in main-paper Table~\ref{tab:recon} under the matched CLIP-identification and SwAV-distance protocol.}
\label{tab:subject_reconstruction}
\end{center}
\end{minipage}
]

\paragraph{Reconstruction analysis.}
The ATM and AVDE profiles show that AVDE's average gain is distributed across subjects, although its size varies by metric and participant. Sub-08 is AVDE's strongest overall row, whereas the best SwAV distance occurs for Sub-10. In the archived ProtoEEG export, AlexNet(2/5) and Inception remain high across participants, with Sub-08 and Sub-10 among the strongest semantic rows. Under the matched main-paper protocol, the ProtoEEG aggregate improves on the AVDE mean by 0.007 PixCorr, 0.024 SSIM, 0.013 AlexNet(2), 0.020 AlexNet(5), 0.019 Inception, and 0.021 CLIP, while reducing SwAV distance by 0.080.

\subsection{Additional Reconstruction Observations}

\paragraph{Low-level variation.}
For ATM, PixCorr ranges from 0.075 to 0.163 and SSIM from 0.301 to 0.363; AVDE reaches 0.188/0.396 on Sub-08. The archived ProtoEEG subject rows range from 0.122 to 0.157 in PixCorr and from 0.246 to 0.273 in SSIM. These values describe participant variation in that export, while panel (c) remains the matched basis for the aggregate comparison in the main paper.

\paragraph{Semantic consistency.}
The archived ProtoEEG rows show comparatively limited variation in AlexNet(2/5) and Inception, although Sub-05 is consistently weaker and Sub-10 is strongest on Inception and CLIPScore. The matched aggregate gains in panel (c) are positive for AlexNet(2), AlexNet(5), Inception, and CLIP, supporting the main-paper conclusion that the prototype code preserves semantic information without claiming a metric-by-metric subject comparison across different evaluator implementations.

\paragraph{Evaluator boundary.}
Panels (a) and (b) retain their original evaluator definitions and are included for subject-level transparency. Panel (c) uses the same seven metric names, directions, formatting, and ProtoEEG aggregate values as main-paper Table~\ref{tab:recon}. In particular, CLIPScore/SwAV-ID in panel (b) must not be read as CLIP identification/SwAV distance in panel (c).

\section{Efficiency and Scope}
\label{app:efficiency}

ProtoEEG has 9.72M trainable parameters, a latency of 0.75\,ms per sample, and a throughput of 92k samples/s on a single A100 (batch size 256, FP16). The EEG encoder accounts for approximately 80\% of latency, prototype assignment and reconstruction for 10\%, and gallery retrieval for the remaining 10\%; the $64\times1024$ codebook is therefore not a computational bottleneck. Matched latency measurements for the baselines remain incomplete, so Table~\ref{tab:appendix_efficiency} reports them as TBD rather than making a comparative speed claim. The EEG-ImageNet experiment is within-dataset retraining rather than cross-dataset transfer, and SI decoding remains substantially harder than SD decoding. Reconstruction conclusions are also tied to the frozen SDXL IP-Adapter and may change with a different generator.

\begin{center}
\centering
\captionof{table}{Inference efficiency. ProtoEEG is measured on a single NVIDIA A100 80\,GB, batch size 256, FP16, averaged over 1000 warm-up and 5000 measurement iterations.}
\label{tab:appendix_efficiency}
\small\setlength{\tabcolsep}{6pt}
\resizebox{\columnwidth}{!}{%
\begin{tabular}{lccc}
\toprule
\textbf{Method} & \textbf{Parameters} & \textbf{Latency (ms)} & \textbf{Throughput (s/s)} \\
\midrule
ProtoEEG    & 9.72M   & 0.75 & 92{,}000 \\
ATM         & $\sim$15M & TBD & TBD \\
NeuroBridge & $\sim$20M & TBD & TBD \\
\bottomrule
\end{tabular}%
}
\end{center}

\end{document}